\documentclass[eat,twocolumn]{jmlr}
   
\usepackage{longtable}

\usepackage{booktabs}
\usepackage[load-configurations=version-1]{siunitx} 

\theorembodyfont{\upshape}
\theoremheaderfont{\scshape}
\theorempostheader{:}
\theoremsep{\newline}

\jmlrvolume{1}
\firstpageno{1}

\jmlryear{2022}
\jmlrworkshop{Machine Learning for Health (ML4H) 2022}

\title[Treatment-RSPN: RSPNs for Sequential Treatment Regimes]{Treatment-RSPN: Recurrent Sum-Product Networks\titlebreak for Sequential Treatment Regimes}

 \author{\Name{Adam Dejl} \Email{adamdejl@mit.edu}\\
 \addr Massachusetts Institute of Technology
 \AND
 \Name{Harsh Deep} \Email{hdeep@mit.edu}\\
 \addr Massachusetts Institute of Technology
 \AND
 \Name{Jonathan Fei} \Email{jyfei@mit.edu}\\
 \addr Massachusetts Institute of Technology
 \AND
 \Name{Ardavan Saeedi} \Email{av.saeedi@gmail.com}\\
 \addr Optum Labs\thanks{Work is not related to the research done at Optum Labs.} 
 \AND
 \Name{Li-wei H. Lehman} \Email{lilehman@mit.edu}\\
 \addr Massachusetts Institute of Technology \\
 \addr MIT-IBM Watson AI Lab
 }
  
\usepackage{booktabs}
\usepackage{multirow}
\usepackage{amsmath}
\usepackage{amssymb}
\usepackage{pdfrender}
\usepackage{pifont}
\usepackage{enumitem}

\begin{document}

\maketitle

\begin{abstract}
Sum-product networks (SPNs) have recently emerged as a novel deep learning architecture enabling highly efficient probabilistic inference. Since their introduction, SPNs have been applied to a wide range of data modalities and extended to time-sequence data. In this paper, we propose a general framework for modelling sequential treatment decision-making behaviour and treatment response using recurrent sum-product networks (RSPNs). Models developed using our framework benefit from the full range of RSPN capabilities, including the abilities to model the full distribution of the data, to seamlessly handle latent variables, missing values and categorical data, and to efficiently perform marginal and conditional inference. Our methodology is complemented by a novel variant of the expectation-maximization algorithm for RSPNs, enabling efficient training of our models. We evaluate our approach on a synthetic dataset as well as real-world data from the MIMIC-IV intensive care unit medical database. Our evaluation demonstrates that our approach can closely match the ground-truth data generation process on synthetic data and achieve results close to neural and probabilistic baselines while using a tractable and interpretable model.
\end{abstract}

\section{Introduction}
\label{sec:intro}
Modelling of the sequential decision-making process of physicians, as well as the patient response to different treatments, has a wide range of useful applications. Transparent and interpretable models of human decision-making behaviour have been proposed as tools for understanding, quantifying and replicating policies in clinical practice \citep{interpole-huyuk, poetree-pace}, while the treatment response models can help to inform clinician's choices when choosing among multiple available therapies \citep{forecasting-treatment-responses-lim}.

Unfortunately, the currently widely used modelling methods suffer from considerable limitations. Neural-based models are highly expressive but do not represent tractable probabilistic distributions and are thus unable to support exact joint, marginal or conditional inference, greatly limiting their versatility. This also means that such models are inherently opaque and incapable of quantifying the uncertainty of their predictions in a principled way. Meanwhile, conventional probabilistic models commonly rely on highly specialised approaches for training and inference, incur high computational costs when evaluating certain probabilistic queries and are often insufficiently expressive to faithfully represent more complex systems.

Sum-product networks (SPNs) \citep{sum-product-networks-poon, darwiche-differential-approach-inference}, a deep learning architecture based on probabilistic circuits, largely address the above limitations, as they are highly expressive while still guaranteeing tractability. Additionally, SPNs are capable of performing fast inference on high treewidth dependencies \citep{vergari-simplifying-structure-learning}, which is considered impossible for other classes of probabilistic models such as Bayesian networks \citep{kwisthout-necessity-bounded-treewidth-bayesian}. Due to these appealing properties, SPNs have been successfully applied in a variety of settings, including face image completion \citep{sum-product-networks-poon}, robot navigation \citep{spn-graph-zheng}, image segmentation \citep{spn-image-segmentation-friesen}, speech modelling \citep{speech-modeling-spn-peharz} or activity recognition in videos \citep{spn-activity-recognition-amer}. SPNs have also been generalized to time-series data of variable length, giving rise to recurrent sum-product networks (RSPNs) \citep{dynamic-spns-melibari}.

In this work, we introduce a probabilistic deep generative approach, Treatment-RSPN, that leverages RSPNs for joint modelling of treatment decision-making and the dynamics of the patient's response to different treatments. As part of our framework, we develop a method for transforming conventional probabilistic graphical models (PGMs), such as dynamic Bayesian networks (DBNs) \citep{dynamic-bayesian-dean}, into RSPNs, allowing us to bootstrap our models with a structure informed by domain knowledge and the specific task. Initializing the models in that way also benefits interpretability, as the random variables from the underlying PGM and the dependencies between them are faithfully reflected in the resulting RSPN. In order to train our models on data, we introduce a variant of the expectation-maximization (EM) algorithm optimizing the parameters of the RSPN and optionally updating its structure to more closely represent the considered distribution.

Treatment-RSPN provides \emph{interpretable} and \emph{tractable} models for sequential treatment regimes in partially observed settings. Additionally, these models benefit from the full range of the appealing properties of SPNs, including the existence of generic algorithms for fast and efficient evaluation of joint, marginal and conditional probabilistic queries, seamless handling of latent variables and a high degree of extensibility offered by structure learning algorithms adapting the network to better match the data. Our code will be made available at \url{https://github.com/ML-Health/treatment-rspn}.

\section{Background}
\label{sec:background}
The models developed using the Treatment-RSPN framework are based on recurrent sum-product networks (RSPNs) \citep{dynamic-spns-melibari}, a generalization of SPNs specifically adapted for modelling time-series data. Each RSPN is composed of a top network, a template network and a bottom network. The template network serves as a generic interconnecting component representing the variables in the intermediate slices of the considered temporal sequences, while the top and bottom networks enable the special-case handling of the variables in the first and last time slices, respectively. The three networks can be stacked together to form a regular, unrolled SPN capable of representing sequences up to a certain finite length $L$. In this process, the top network of the RSPN is taken as the root component, the template network is repeated $L - 2$ times and interconnected with other components via dedicated placeholder nodes, and the bottom network is added as the terminating component with no further outgoing connections.

The unrolled SPN is a rooted directed acyclic graph with leaf nodes representing base distributions over random variables (such as Categorical or Gaussian) and internal nodes representing sums and products of probability functions of their children. Each edge leading from a sum node to a child node is associated with a non-negative parameter indicating the weight of the corresponding child. The weights of the edges originating at a single sum node must be normalized. A key component of the SPN inference and training methods is the SPN evaluation. Given a (partial or full) assignment to the variables in the scope of an SPN $\mathbf{x}$, the value $S(\mathbf{x})$ of an SPN can be computed as follows:
\scriptsize
\begin{equation*}
    S(\mathbf{x}) =
    \begin{cases}
        P_S(X = \mathbf{x}) & \text{if $S$ is a leaf} \\
        \sum_{C_i \in children(S)}w_i C_i(\mathbf{x}) & \text{if $S$ is a sum} \\
        \prod_{C_i \in children(S)}C_i(\mathbf(x)) & \text{if $S$ is a product}
    \end{cases}
\end{equation*}
\normalsize
The value $S(\mathbf{x})$ corresponds to the probability of the assignment $\mathbf{x}$ under the distribution modelled by the given SPN \citep{sum-product-networks-poon}.

The possibility of evaluating an SPN on a partial variables assignment provides a straightforward way for performing marginal inference, as the variables not included in an assignment are being effectively marginalized over during the computation of the probability of the assignment. This can be used for handling latent variables and missing values during both training and inference, as well as an efficient evaluation of conditional queries, leveraging the fact that $P(\mathbf{q}|\mathbf{e}) = \frac{P(\mathbf{q}, \mathbf{e})}{P(\mathbf{e})}$.

\section{Treatment-RSPN}

\subsection{Treatment-RSPN Initialization}
\label{sec:initialization}
The first step of developing a model within our framework is the initialization of its structure. This step is of particular importance for RSPNs, as their structure directly affects the assumed dependencies between the variables in the scope of the model. In general, there are multiple ways in which the structure can be initialized, including learning the structure directly from the data using one of the structure learning algorithms (such as local search \citep{dynamic-spns-melibari} or oSLRAU \citep{online-structure-learning-kalra}), crafting the structure manually or basing it on a known probabilistic model.

In this work, we opt for the last approach, though we may have chosen one of the other options without changing the rest of our methodology. Specifically, we base the initial structure of our RSPN models on an input-output hidden Markov model (IOHMM) \citep{iohmm-neurips, iohmm-ieee}. IOHMMs model sequences composed of inputs $u_1, ..., u_n$, unobserved (latent) states $z_1, ..., z_n$ and outputs $x_1, ..., x_n$. This makes them highly suited for modelling sequential treatment regimes, as the treatment actions at each time step can be considered as inputs, while the patient physiological variables or test results can be considered as outputs. Additionally, the dependencies between the variables in an IOHMM model are rather simple and intuitive, leading to a more interpretable model.

In order to transform an IOHMM into an RSPN structure, we first consider an unrolled Bayesian network (BN) representation of the model over a fixed number of time slices. We then convert this network into an SPN structure, gradually constructing layers for the individual variables.

The construction procedure processes the variables in the BN representation of the model in topological order, maintaining metadata about their dependent child variables. When processing each variable, our procedure first constructs an SPN sum layer with one node for each possible assignment of this variable as well as variables whose children have not yet been fully processed. This sum layer is then extended with an additional structure, which may either be composed of variable indicators or product nodes connected to such indicators. The product nodes are used in cases in which the currently processed variable has children in the network and facilitate conditioning on this variable during the SPN evaluation. The weights on the edges between the sum nodes and product nodes are determined by the conditional probability distributions modelled by the Bayesian network. We give a detailed algorithmic description of this step of the initialization in the appendix \ref{apd:bn-to-spn}.

After the initial conversion, the repeated portions of the resulting SPN structure modelling the variables in the intermediate time slices can be straightforwardly extracted into a template network, while the structures for the first and the last time slices can be turned into the top and bottom networks, respectively. This results in the initial form of the RSPN. Note that our transformation is fully general and could be applied to an arbitrary model as long as it is expressible as a (dynamic) Bayesian network. We provide an illustrative example of an IOHMM and its associated RSPN representation in appendix~\ref{apd:figs}.

\subsection{Treatment-RSPN Training}
After deriving the initial structure for the RSPN model, we need to learn its parameters based on data. We opt to use an adapted version of the expectation-maximization (EM) algorithm. Compared to other formulations in the literature \citep{sum-product-networks-poon, latent-variable-spn-peharz}, our variant of the EM algorithm does not require the use of differentiation and is straightforwardly applicable to RSPNs in addition to regular SPNs. Additionally, our approach is capable of adjusting the structure of the learned RSPN so that it more closely matches the distribution of the training data.

Our EM algorithm first unrolls the RSPN for a fixed number of time steps required to fit the given training data, internally assigning identical identifiers to the nodes repeated in the unrolled SPN. This ensures that the parameters of these nodes are updated jointly and do not diverge during the training process. The algorithm then alternates between the expectation and maximization steps. In an E-step, the procedure computes the likelihood of each data point "reaching" each node, i.e. the probability of this point having been generated by the node. In an M-step, the algorithm updates the weights on the edges leading from the sum nodes as well as the internal parameters of the leaves according to these likelihoods. Optionally, the algorithm can also replace the SPN leaves with a more complex learned structure by clustering the data points weighted by their likelihoods at the given leaf and creating a new sum node with multiple leaves modelling the clusters. We give a detailed description of our EM algorithm in the appendix \ref{apd:em}.

\subsection{Interpretability}
\label{sec:interpretability}
\begin{table}[!tb]
     \scriptsize
     \centering
     \begin{center}
     \begin{tabular}{ c c } 
        \toprule
        \multirow{2}{*}{\textbf{Model}} & \textbf{Log-likelihood ($\uparrow$)} \\ 
        \cmidrule(r){2-2}
        & \textbf{Mean} $\pm$ \textbf{SD} \\ 
        \toprule
        Reference IOHMM & -18842.82 $\pm$ 131.44\\
        \midrule
        Treatment-RSPN & -18885.07 $\pm$ 129.20 \\
        \bottomrule
     \end{tabular}
     \end{center}   \caption{Log-likelihood of the synthetic test data under the reference IOHMM model used for their generation and the learned Treatment-RSPN model}
     \label{fig:synthetic-iohmm}
\end{table}
We argue that our model is more interpretable compared to other (especially neural-based) methods due to its increased transparency associated with tractability as well as its basis in a human-intuitive IOHMM model with clearly defined dependencies between the involved variables. The tractability of Treatment-RSPN allows its probing with arbitrary joint, marginal or conditional queries, which may be used to answer questions about the learned probabilistic beliefs of the model.

Additionally, the latent, hidden states introduced by the underlying IOHMM model can also be used to produce explanations of the model behaviour. For example, it is possible to inspect the latent states most commonly associated with certain predictions and to visualize their associated emission distributions, which can provide insight about typical clinical states of patients for which the model suggests a particular treatment. We give examples and more details on the interpretability of our approach in the appendix \ref{apd:interpretability}.

\section{Experiments}
\label{sec:experiments}
We evaluate the performance of the Treatment-RSPN models on a synthetically generated dataset as well as real-world data from the MIMIC-IV intensive care unit medical database \citep{mimic-iv-johnson, physionet-goldberger}.
\begin{table*}[!tb]
     \scriptsize
     \centering
     \begin{center}
     \begin{tabular}{ c c c c c c } 
        \toprule
        \multirow{2}{*}{\textbf{Model}}  & \textbf{AUROC ($\uparrow$)} & \textbf{F1 macro ($\uparrow$)} & \textbf{Brier score ($\downarrow$)} \\ 
        \cmidrule(r){2-2}
        \cmidrule(r){3-3}
        \cmidrule(r){4-4}
       & \textbf{Mean} $\pm$ \textbf{SD} & \textbf{Mean} $\pm$ \textbf{SD} & \textbf{Mean} $\pm$ \textbf{SD} \\ 
        \toprule
        Interpole & 0.88 $\pm$ 0.020 &  0.87 $\pm$ 0.023& 0.042 $\pm$ 0.021 \\
        PO-MB-IL  & 0.58 $\pm$ 0.088  &  0.49 $\pm$ 0.010 & 0.108 $\pm$ 0.065 \\
        PO-IRL  & 0.57 $\pm$ 0.016 &  0.44 $\pm$ 0.020 & 0.437 $\pm$ 0.003 \\
        Off. PO-IRL & 0.57 $\pm$ 0.054 & 0.49 $\pm$ 0.009 & 0.484 $\pm$ 0.003 \\
        LSTM  & 0.85 $\pm$ 0.030 & 0.85 $\pm$ 0.023 & 0.049 $\pm$ 0.021 \\
        Predict most common & 0.50 $\pm$ 0.000 &  0.49 $\pm$ 0.009 & 0.098 $\pm$ 0.061 \\
        \midrule
        Treatment-RSPN  & \textbf{0.91 $\pm$ 0.010}  &  0.86 $\pm$ 0.008 & 0.045 $\pm$ 0.024 \\
        \bottomrule
     \end{tabular}
     \end{center}   \caption{Classification and calibration performance of the evaluated models on the one-step-ahead treatment action prediction task using the MIMIC-IV sepsis dataset. The values highlighted in bold mark significantly better results at $0.05$ confidence level.}
     \label{fig:vaso-iohmm-eval-action}
\end{table*}
\begin{table*}[!tb]
     \scriptsize
     \centering
     \begin{center}
     \begin{tabular}{ c c c c c c c c c  } 
        \toprule
        \multirow{2}{*}{\textbf{Model}} & \textbf{HR RMSE ($\downarrow$)} & \textbf{MBP RMSE ($\downarrow$)} & \textbf{RR RMSE ($\downarrow$)} & \textbf{AVG RMSE ($\downarrow$)} \\ 
        \cmidrule(r){2-2}
        \cmidrule(r){3-3}
        \cmidrule(r){4-4}
        \cmidrule(r){5-5}
        & \textbf{Mean} $\pm$ \textbf{SD} & \textbf{Mean} $\pm$ \textbf{SD} & \textbf{Mean} $\pm$ \textbf{SD} & \textbf{Mean} $\pm$ \textbf{SD} \\ 
        \toprule
        LSTM & 10.81 $\pm$ 0.525 & 14.57 $\pm$ 0.938 & 5.35 $\pm$ 0.137 & 10.24    
        $\pm$ 0.487 \\
        \midrule
        Treatment-RSPN & \textbf{9.54 $\pm$ 0.267} & \textbf{13.09 $\pm$ 0.453} & \textbf{4.46 $\pm$ 0.209} & \textbf{9.03 $\pm$ 0.286} \\
        \bottomrule
     \end{tabular}
     \end{center}
     \caption{Regression performance of the evaluated models on the one-step ahead treatment response prediction task using the MIMIC-IV sepsis dataset. The values highlighted in bold mark significantly better results at $0.05$ confidence level.}
     \label{fig:vaso-iohmm-eval-reponse-continuous}
\end{table*}

In the synthetic data experiment, we test whether the Treatment-RSPN can closely match the parameters of a ground truth data generation process based on an IOHMM with two inputs, two states and four possible categorical observations. As shown in table \ref{fig:synthetic-iohmm}, the log-likelihood of the Treatment-RSPN differs from the log-likelihood of the ground-truth model by less than $0.25\%$, indicating a near-perfect match.

In the first MIMIC-IV experiment, we apply Treatment-RSPN to one-step-ahead treatment action prediction of the administration of vasopressors, with discretized minimum blood pressure and the total volume of intravenous fluids received by the patient serving as covariates. The task is set up as a binary classification problem, with the target values indicating whether any vasopressors were administered at the corresponding time step. We compare the performance of our model to baselines from \citep{interpole-huyuk}, as well as a basic model always predicting the most common treatment action. The results of the experiment are captured in table \ref{fig:vaso-iohmm-eval-action}. The Treatment-RSPN achieved a statistically significantly improved result over all baselines in the AUROC score and is on par with Interpole (the best performing baseline) in terms of F1 macro and Brier score.

In the second MIMIC-IV experiment, we evaluate the performance of Treatment-RSPN on one-step-ahead continuous treatment response prediction, using vasopressors and intravenous fluids administration indicators as inputs and heart rate, minimum blood pressure and respiratory rate as the predicted treatment outcome variables. This task can thus be seen as a regression task with three predicted variables. The results of Treatment-RSPN and an LSTM baseline \citep{lstm-hochreiter} on this task are shown in table \ref{fig:vaso-iohmm-eval-reponse-continuous}. We can see that the Treatment-RSPN model significantly outperforms the LSTM baseline.

In both MIMIC-IV experiments, the Treatment-RSPN achieved a performance competitive with the baselines, demonstrating the promise of the method.

\section{Conclusion and Future Work}
In this paper, we introduced a framework for tractable and interpretable modelling of clinical decision-making and treatment response mechanisms using recurrent-sum-product networks. We discussed the process for constructing the initial network structure for the RSPN models, as well as an adapted version of the EM algorithm suitable for their training. In the future, we would like to extend the structure-learning capabilities of our training algorithm and further explore how the interpretability and tractability of our models could be used to provide more informative predictions. We would also like to explore the potential use of Treatment-RSPN for simulation.

\acks{The first author was supported by the Turing Scheme. L Lehman was in part funded by MIT-IBM Watson AI Lab. We thank the anonymous reviewers for their helpful and constructive suggestions.}

\bibliography{refs}

\appendix
\section{Treatment-RSPN Overview Diagram}
\begin{figure*}[!p]
  \floatconts
  {fig:treatment-rspn}
  {\caption{Overview of the Treatment-RSPN framework. Treatment-RSPN is initialised based on a probabilistic model (e.g. IOHMM) and trained on sequential treatment data using our expectation maximization algorithm. The resulting model can be applied to different tasks, such as treatment action prediction and treatment response prediction.}}
  {\includegraphics[width=1.0\linewidth]{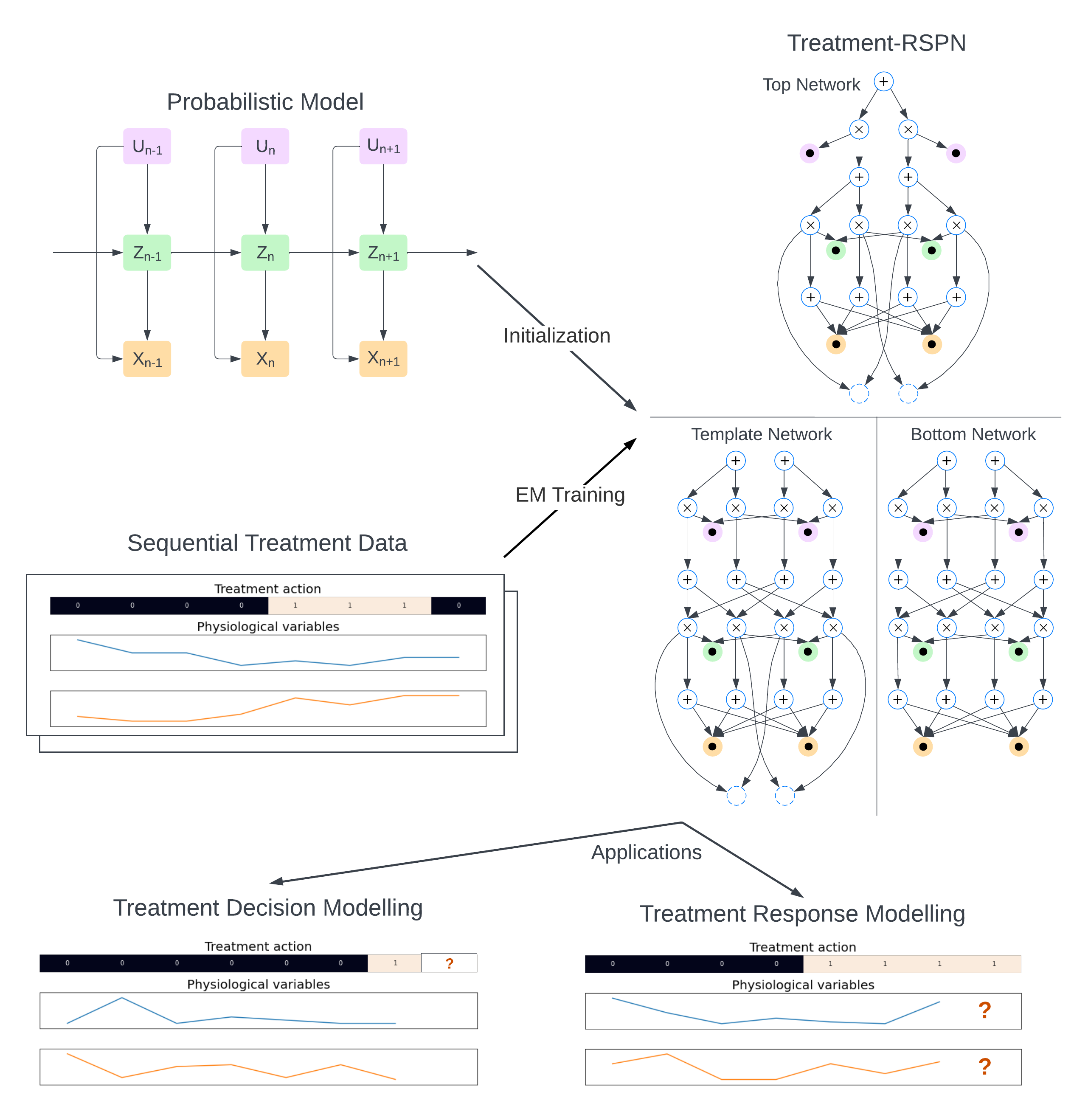}}
\end{figure*}
An overview diagram illustrating our approach is provided in figure \ref{fig:treatment-rspn}.

\section{Conversion Procedure}
We give our procedure for converting a Bayesian network into an SPN in algorithm \ref{alg:bn-to-spn}.
\label{apd:bn-to-spn}
\begin{algorithm}[p]
\smaller
\floatconts
{alg:bn-to-spn}
{\caption{\smaller BN to SPN conversion procedure}}
{
Given a Bayesian network $N$ to convert to an SPN
\begin{enumerate*}
  \item Set $V$s to be the variables (nodes) in $N$ in the topological order.
  \item Set $c = \emptyset$ to be the set of variables currently conditioned on.
  \item Set $d = \{\}$ to be a map of unresolved variable dependencies (mapping from parent variables to lists of child variables).
  \item For $V$ in $V$s:
  \begin{enumerate*}
    \item Construct a layer of sum nodes $s$, one for each possible assignment $\textbf{x}^s_i$ of variables in $c$ or a single sum node if $c$ is empty. Attach the layer to the previous product layer $p$ if any, with edges leading between nodes $n^p_j$ and $n^s_i$ whenever $\textbf{x}^p_j$ and $\textbf{x}^s_i$ for these nodes are compatible.
    \item If $V$ has children in $N$:
    \begin{enumerate}[label=\alph*.]
      \item Add $V$ to $c$.
      \item Remove all occurrences of $V$ from $d$, remove any variables with empty child list in $d$ from $c$.
      \item Add $V \mapsto \text{children}_N(V)$ to $d$.
      \item Add a product layer $p$ with one node for each possible assignment $\textbf{x}^p_j$ of variables in $c$, each connected to an indicator leaf for $V = v_j$ where $v_j$ is the value of $V$ in $\textbf{x}^p_j$. Connect the product layer to the previous sum layer $s$, with edges leading between nodes $n^s_i$ and $n^p_j$ whenever $\textbf{x}^s_i$ and $\textbf{x}^p_j$ for these nodes are compatible and edge weights determined by $P_N(V = v_j|\textbf{x}^s_i)$.
    \end{enumerate}
    Otherwise:
    \begin{enumerate}[label=\alph*.]
      \item Construct a layer of leaf nodes for $V$ with distributions and connections to the previous layer $l$ representing the distribution $P_N(V|\textbf{x}^l_i)$ for each node $n^l_i$ with assignment $\textbf{x}^l_i$ in $l$.
    \end{enumerate}
  \end{enumerate*}
\end{enumerate*}
}
\end{algorithm}

\section{EM Algorithm}
\label{apd:em}
We detail our variant of the RSPN expectation maximization procedure in algorithm \ref{alg:em}.

\begin{algorithm}[p]
\smaller
\floatconts
{alg:em}
{\caption{\smaller RSPN expectation maximization}}
{
Given an RSPN $R$ and the training dataset $D$
\begin{enumerate*}
  \item Unroll $R$ into a regular SPN $S$ matching the length of the sequences in $D$ and using shared node identifiers for repeated nodes.
  \item Evaluate $S$ on $D$ using procedure described in appendix \ref{sec:background}, store the value computed at each node.
  \item Set $pl = \{(\text{nil}, \text{root}(R)) \mapsto \mathbf{1}\}$ to be the map of data point likelihoods passed to each node by its parents. The root node has no parents, and is thus passed likelihood of $1$ for each point.
  \item Set $l = \{(\text{id}(p), \text{id}(c)) \mapsto 0\}$ for all pairs of parents and children $(p, c)$ in $S$. $l$ stores the aggregated likelihoods of data points reaching nodes from their parents.
  \item Set $lp, lw = \{\}, \{\}$ to store data points and their likelihoods at each leaf.
  \item For $n$ in nodes($S$), processed from top to bottom:
  \begin{enumerate*}
    \item Set $dl = \sum_{p \in \text{parents($n$)}} pl\text{[$p$, $n$]}$ to be the likelihoods of reaching the current node for each data point, passed to $n$ by its parents.
    \item Increment $l$[id($p$), id($n$)] by $\sum_{d \in D}$ $pl$[$p$, $n$, $d$] for all $p$ in parents($n$).
    \item If $n$ is a sum:
    \begin{enumerate}[label=\alph*.]
        \item Set $cl = \{c \mapsto$ weight($n$, $c$) $\times$ value($c$)$\}$ for all $c$ in children($n$) to store the base likelihood of each child.
        \item Normalize the likelihoods of the children in $cl$
        \item Set $pl$[$n$, $c$]$ = cl$[$c$] $\times$ $dl$
    \end{enumerate}
    If $n$ is a product:
    \begin{enumerate}[label=\alph*.]
        \item Set $pl$[$n$, $c\text{]} = dl$ for all $c$ in children($n$)
    \end{enumerate}
    If $n$ is a leaf node:
    \begin{enumerate}[label=\alph*.]
        \item Store the current data $D$ and their likelihoods at this node $dl$ in $lp$[$n$] and $lw$[$n$].
    \end{enumerate}
  \end{enumerate*}
  \item Update the weights of each edge between a sum node $s$ and its child node $c$ to $\frac{l\text{[id(}s\text{)}, \text{id(}c\text{}]}{\sum_{c' \in \text{children(}s\text{)}}l\text{[id(}s\text{)}, \text{id(}c'\text{)]}}$
  \item Update the distributions in leaves according to the weighted data in $lp$ and $lw$ or cluster the weighted data and introduce a sub-structure to model the mixture.
\end{enumerate*}
}
\end{algorithm}

\section{IOHMM and Treatment-RSPN Diagrams}
\label{apd:figs}
To gain more intuition about the transformation process and the structure of the resulting Treatment-RSPN, consider the example IOHMM in figure \ref{fig:iohmm} and the corresponding RSPN obtained using our conversion procedure in figure \ref{fig:rspn}. There are several key observations to note. First, notice how the product nodes are naturally used to model conditioning on a variable represented by leaves attached to them. This is exploited by the transformation algorithm, which constructs a product layer for each variable with children in the underlying BN. Second, note how the converging connections from multiple product nodes effectively ``cancel'' conditioning on a variable. For example, the edges from the product nodes for the assignments $u_1 = 0, z_1 = 0$ and $u_1 = 1, z_1 = 0$ leading into a single node stop the conditioning on the variable $U_1$. Finally, note that the probabilities from the conditional probability tables associated with the IOHMM model are repeated several times in the network. During the training of our model, we assign identical parameters to the nodes surrounding the edges with these probabilities, which ensures that they are updated in unison.

\begin{figure*}[!p]
  \floatconts
  {fig:iohmm}
  {\caption{A simple IOHMM with discrete variables for the inputs, latent states and observations.}}
  {\includegraphics[width=0.47\linewidth]{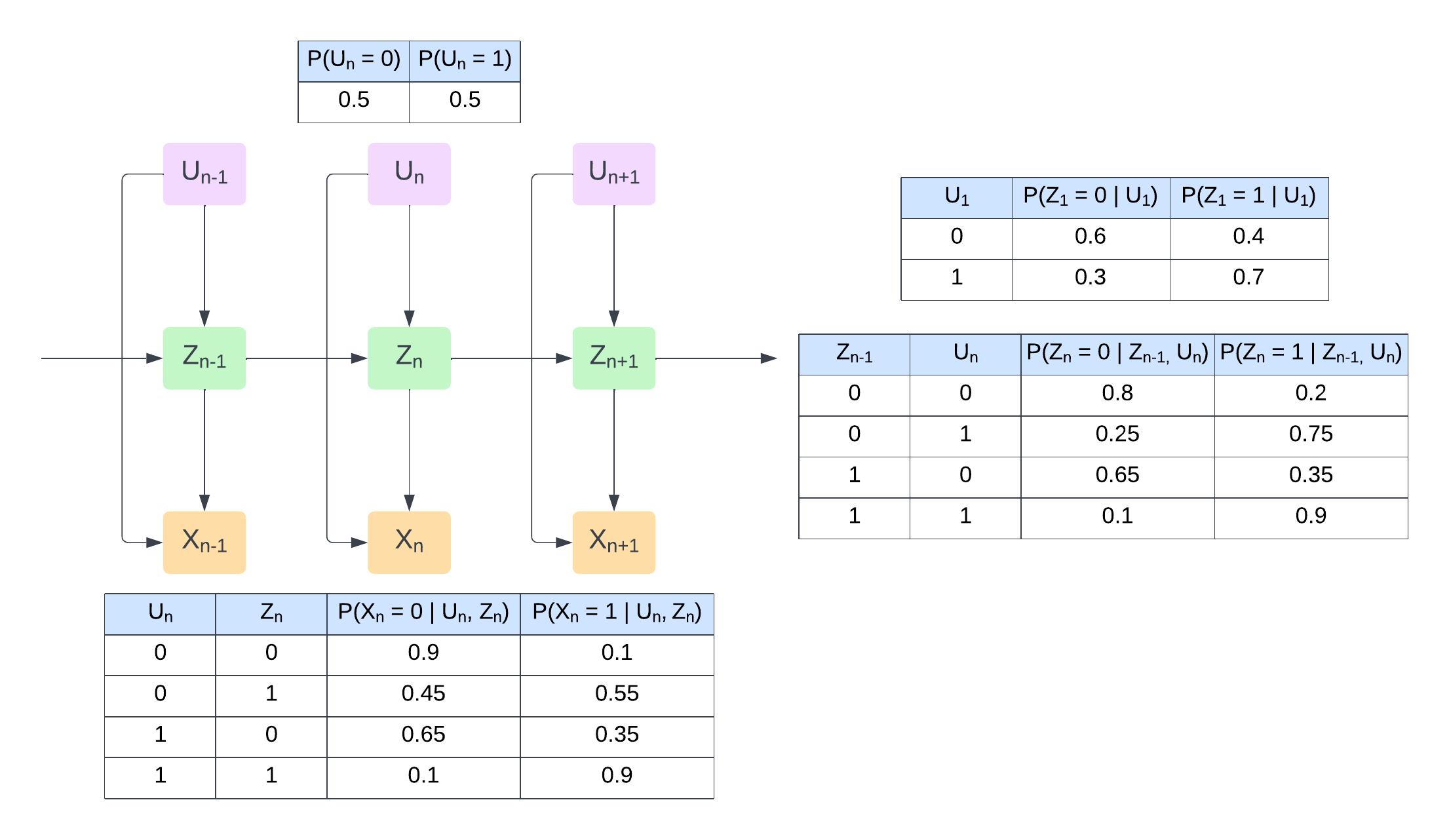}}
\end{figure*}

\begin{figure*}[!p]
  \floatconts
  {fig:rspn}
  {\caption{An RSPN corresponding to the IOHMM model shown in figure \ref{fig:iohmm}. This model can be obtained by running our initialization procedure described in section \ref{sec:initialization}.}}
  {\includegraphics[width=0.5\linewidth]{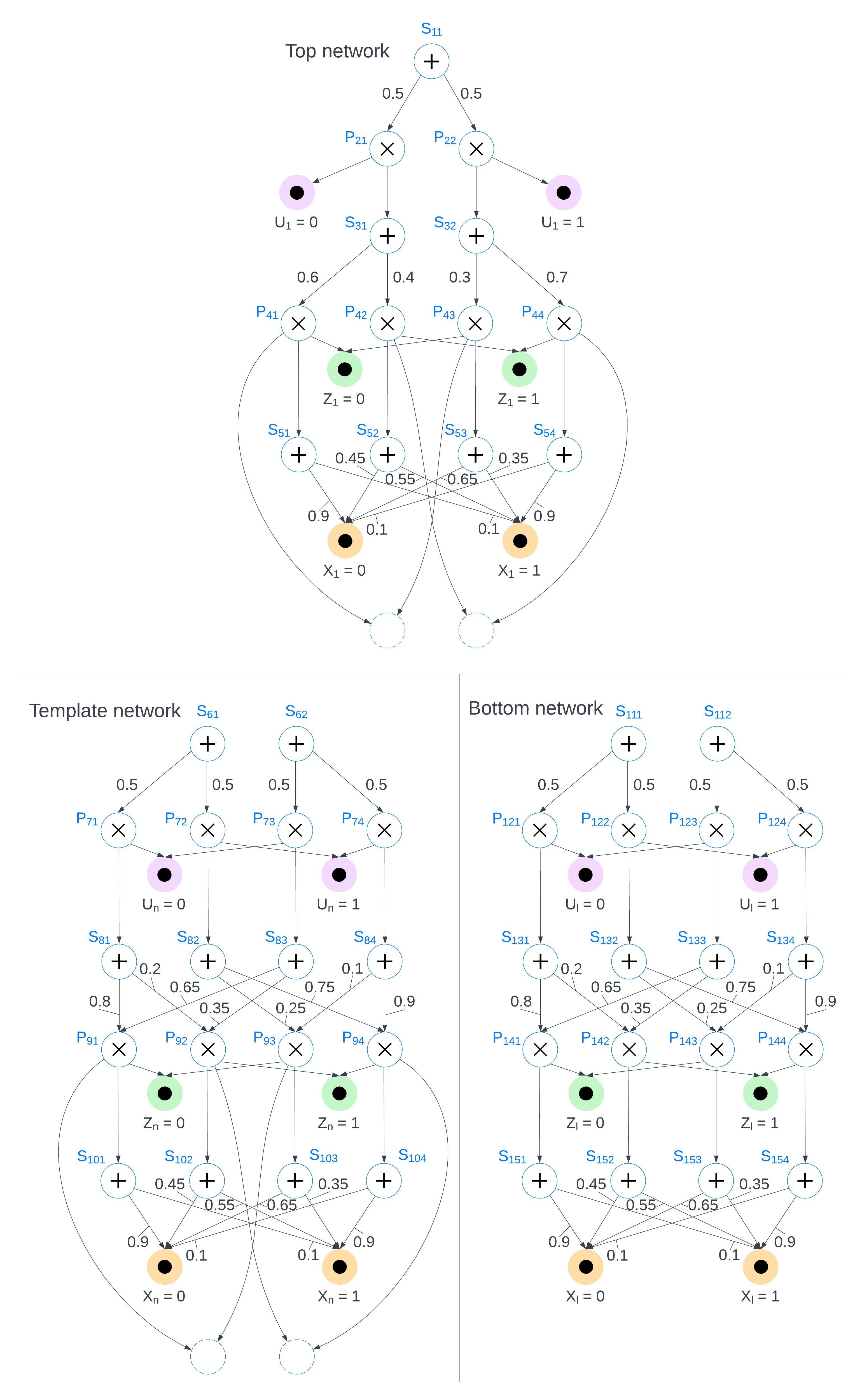}}
\end{figure*}

\section{Interpretability}
\begin{figure*}[!htbp]
    \floatconts
    {fig:interpret}
    {\caption{States with the highest likelihoods before a negative/positive vasopressor treatment}}
     {
        \subfigure[States for negative vasopressor treatment]{\label{fig:states-no-vaso}
            \includegraphics[width=0.48\textwidth]{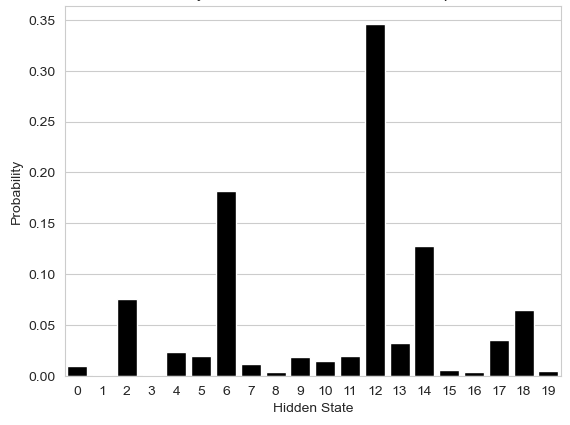}}
        \hfill
        \subfigure[States for positive vasopressor treatment]{\label{fig:states-vaso}
            \includegraphics[width=0.48\textwidth]{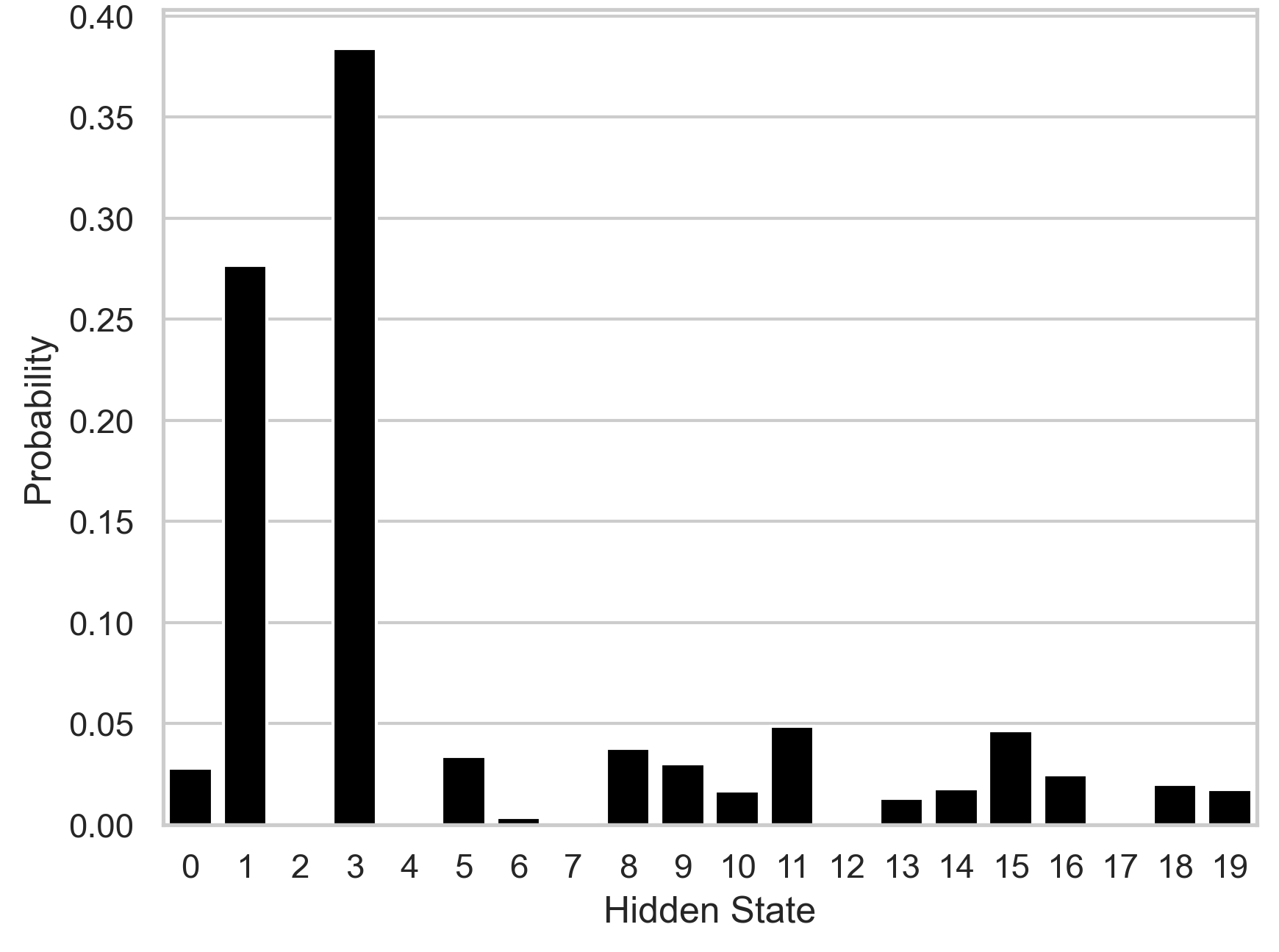}}
     }
\end{figure*}
\begin{figure*}[!htbp]
    \floatconts
    {fig:state_interpret}
    {\caption{Emission distributions for selected states with the highest probabilities before positive vasopressor treatment (states 1 and 3) and negative vasopressor treatment (state 12)}}
    {
        \subfigure[State 1]{\includegraphics[scale=0.32]{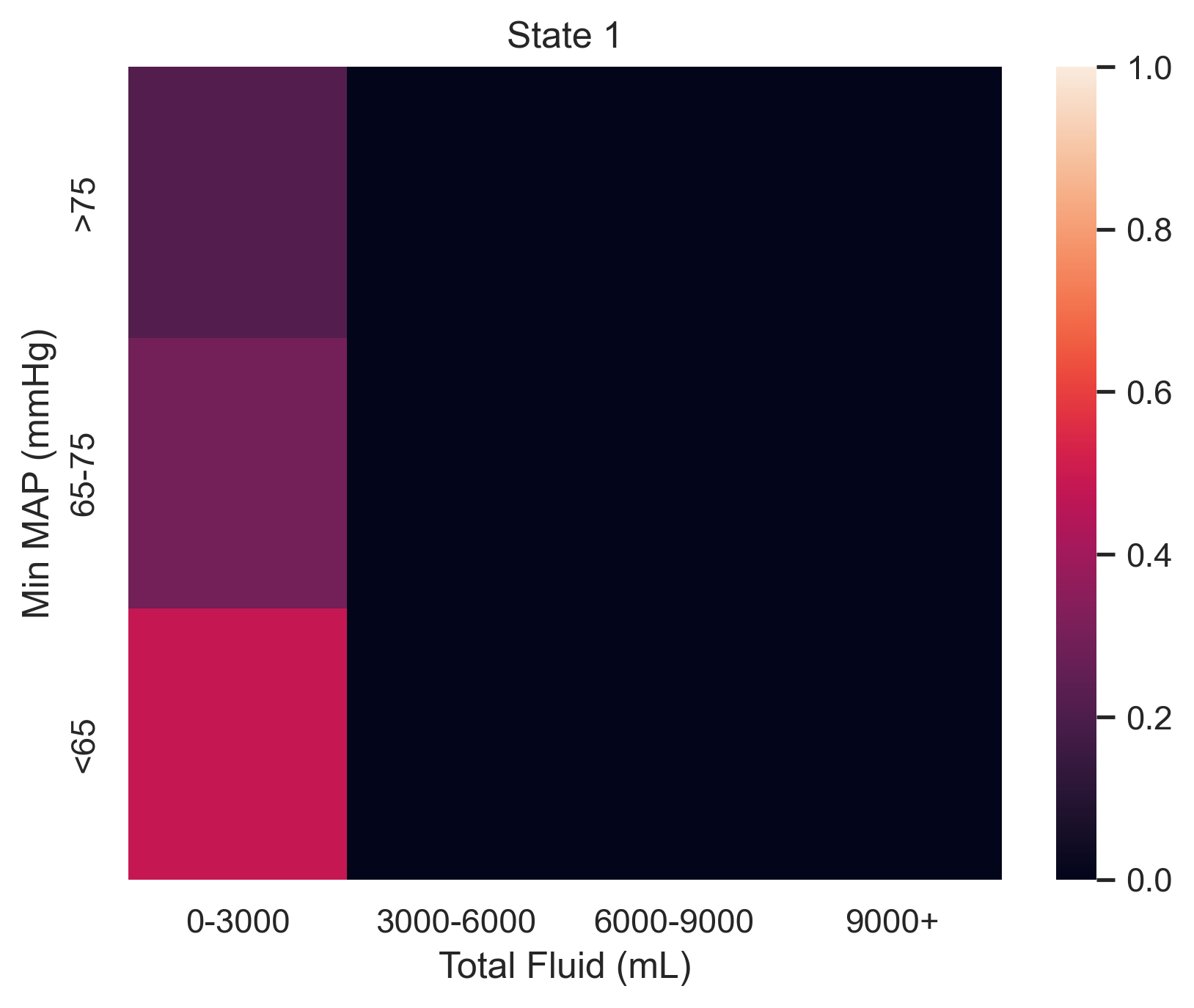}}
        \hfill
        \subfigure[State 3]{\includegraphics[scale=0.32]{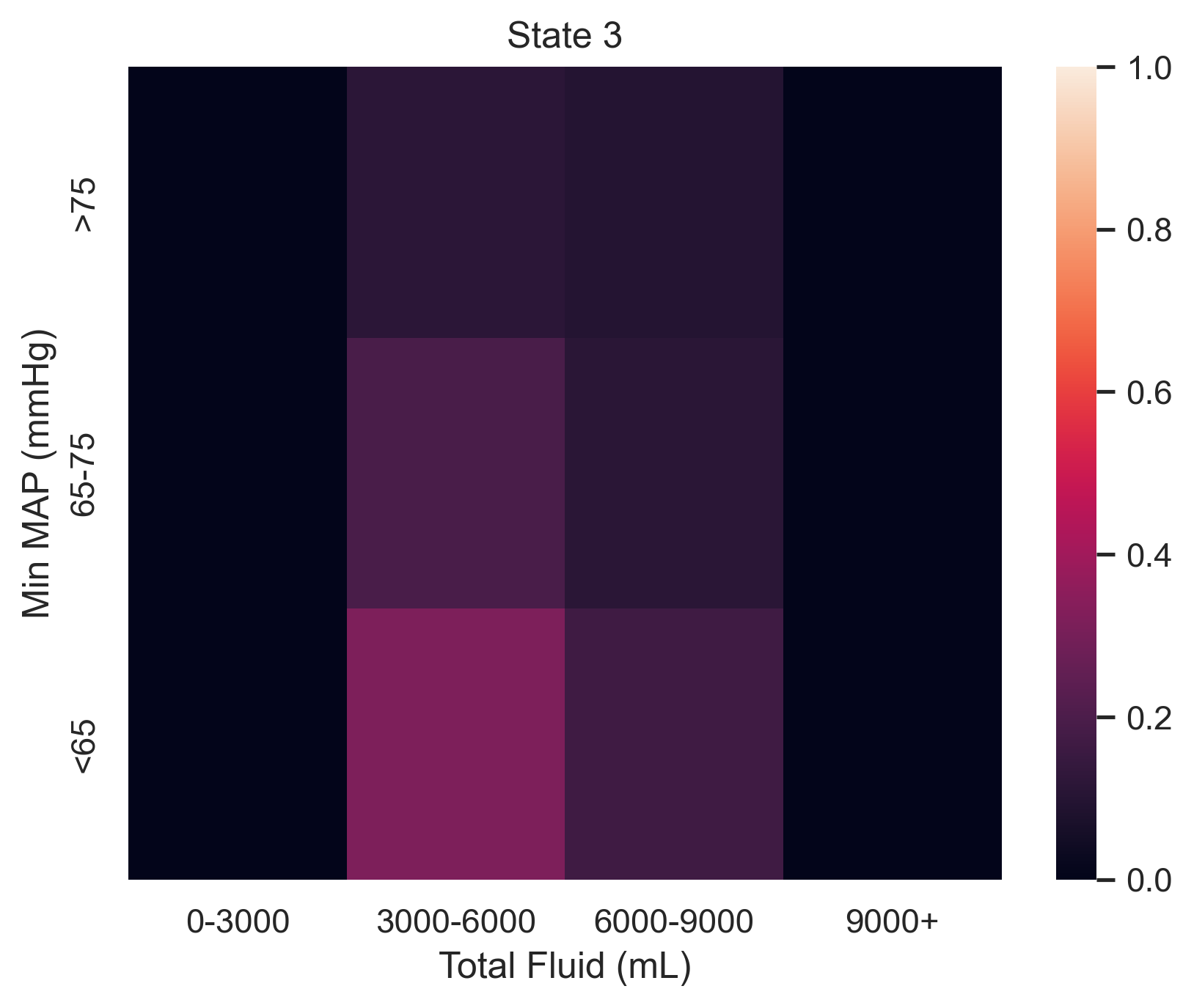}}
        \hfill
        \subfigure[State 12]{\includegraphics[scale=0.32]{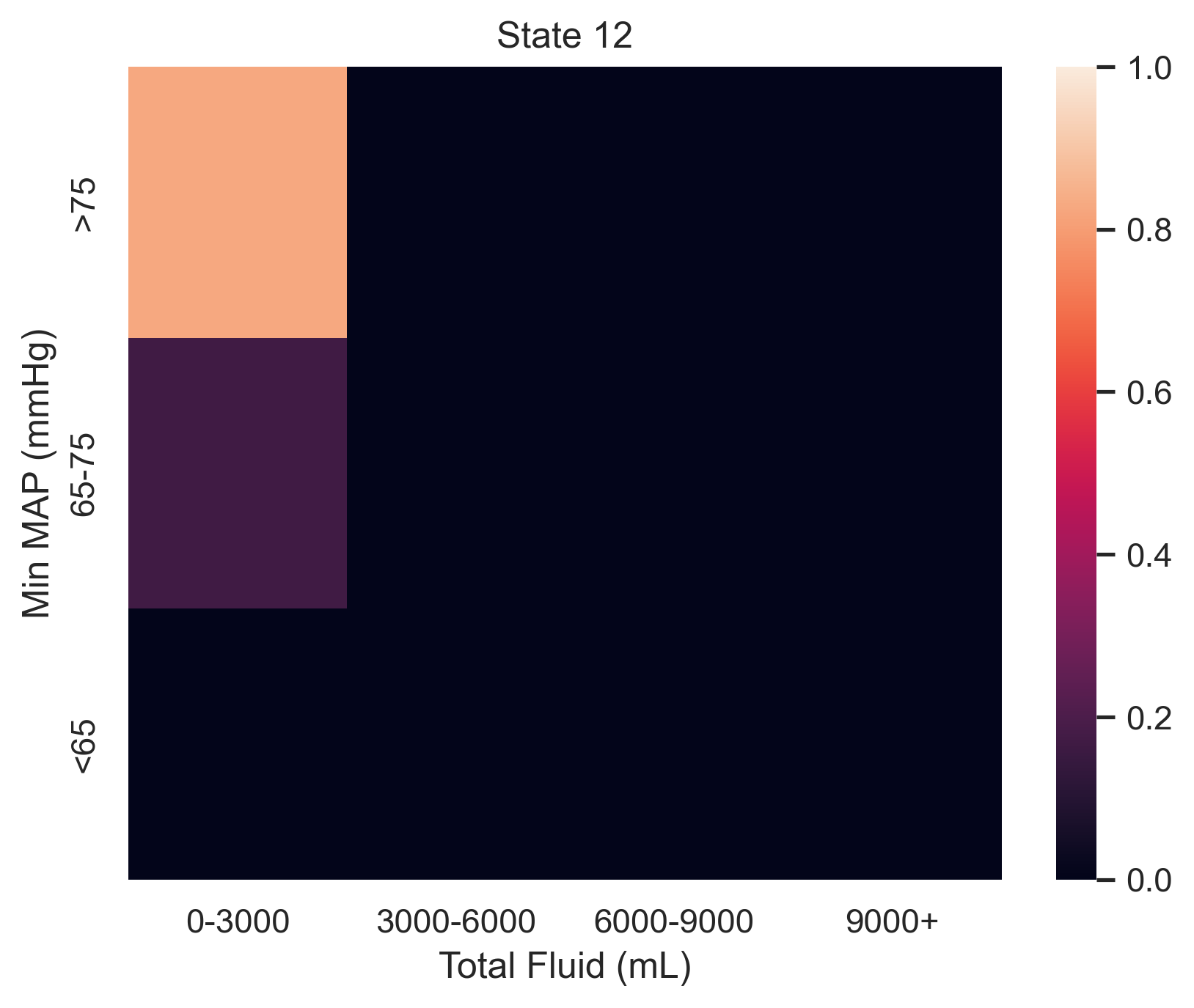}}
    }
\end{figure*}

\label{apd:interpretability}
In section \ref{sec:interpretability} of the main text, we argued that Treatment-RSPN is more interpretable compared to the other approaches due to its probabilistic nature, tractability and basis in a known probabilistic graphical model with intuitive dependencies between the variables. Here, we show an illustrative example of how these properties could be practically utilized to gain insight into the learned beliefs of the model. In our analysis, we focus on the Treatment-RSPN model used for the treatment action prediction experiment.

As a first step, we determine the most likely states to occur immediately before the positive or negative vasopressor treatment action. This is done by computing the likelihoods of the different states at each time step of each temporal sequence in the test set and aggregating them appropriately. The results are visualized in figure \ref{fig:interpret}, showing that the most likely state before a negative vasopressor treatment is state number 12, while the most likely states before a positive vasopressor treatment are states number 1 and 3.

In order to gain further insight into the characteristics of these states, we can examine the emission distributions associated with them. These are visualized in figure \ref{fig:state_interpret}. Each heatmap depicts the probabilities for the observed outcomes in the current time step based on the hidden state. Both states 1 and 3 occur at timesteps where the observed blood pressure is low. This corresponds with the expectations, as clinicians are likely to begin vasopressor therapy for urgently hypotensive patients. It thus seems that the beliefs of the model are clinically meaningful, as it has learned that vasopressors must be administered after low blood pressure. In comparison, patients in state 12, which is predictive of negative vasopressor treatment, are much more likely to have higher blood pressure, indicating that the therapy is not needed.

In addition to the above observations, we can also note that the model apparently learned to distinguish between patients who have been administered a larger volume of intravenous fluids (state 3) and patients who have not received much of these fluids (state 1). This may potentially be helpful for determining the point at which the vasopressor treatment is more likely to cease.

\section{Discrete Treatment Response Prediction Experiment}
In addition to the experiments described in section \ref{sec:experiments}, we also evaluated our model on the ``mirror'' task to MIMIC-IV experiment one, in which the model performs a one-step-ahead prediction of the discretized covariates (minimum blood pressure and the total volume of intravenous fluids). The results for this task are shown in table \ref{fig:vaso-iohmm-eval-reponse-discrete}.

\begin{table*}[!htp]
     \small
     \centering
     \begin{center}
     \begin{tabular}{ c c c c c c }
        \toprule
        \multirow{2}{*}{\textbf{Model}} & \textbf{F1 weighted ($\uparrow$)} & \textbf{AUROC weighted ($\uparrow$)} & \textbf{Brier score ($\downarrow$)} \\ 
        \cmidrule(r){2-2}
        \cmidrule(r){3-3}
        \cmidrule(r){4-4}
        \cmidrule(r){5-5}
        & \textbf{Mean} $\pm$ \textbf{SD} & \textbf{Mean} $\pm$ \textbf{SD} & \textbf{Mean} $\pm$ \textbf{SD} \\ 
        \toprule
        LSTM & 0.64 $\pm$ 0.054 & 0.93 $\pm$ 0.012 & 0.49 $\pm$ 0.058 \\
        Predict most common & 0.10 $\pm$ 0.062 & 0.50 $\pm$ 0.000 & 0.87 $\pm$ 0.031 \\
        \midrule
        Treatment-RSPN & 0.60 $\pm$ 0.059 & 0.93 $\pm$ 0.011 & 0.52 $\pm$ 0.062 \\
        \bottomrule
     \end{tabular}
     \end{center}
     \caption{Classification performance of the evaluated models on the one-step ahead treatment response prediction task using the MIMIC-IV sepsis dataset}
     \label{fig:vaso-iohmm-eval-reponse-discrete}
\end{table*}

\end{document}